# Synthesized Annotation Guidelines are Knowledge-Lite Boosters for Clinical Information Extraction

**Enshuo Hsu**[1,3], MS; **Martin Ugbala, BS**[2]; **Krishna Kumar Kookal, MS**[2]; **Kawtar Zouaidi**[2], Dr. med. dent, MPH; **Nicholas L Rider**[4], DO; **Muhammad F Walji**[1,2], PhD; **Kirk Roberts**[1], PhD

[1]McWilliams School of Biomedical Informatics, University of Texas Health Science Center at Houston, Houston, Texas, USA

[2]School of Dentistry, University of Texas Health Science Center at Houston, Houston, Texas, USA

[3]Enterprise Development and Integration, University of Texas MD Anderson Cancer Center, Houston, Texas, USA

[4]Virginia Tech Carilion School of Medicine, Department of Health Systems & Implementation Science, Roanoke, Virginia, USA

**Corresponding author: Kirk Roberts, kirk.roberts@uth.tmc.edu**

## ABSTRACT

Generative information extraction using large language models, particularly through few-shot learning, has become a popular method. Recent studies indicate that providing a detailed, human-readable guideline—similar to the annotation guidelines traditionally used for training human annotators—can significantly improve performance. However, constructing these guidelines is both labor- and knowledge-intensive. Additionally, the definitions are often tailored to meet specific needs, making them highly task-specific and often non-reusable. Handling these subtle differences requires considerable effort and attention to detail. In this study, we propose a self-improving method that harvests the knowledge summarization and text generation capacity of LLMs to synthesize annotation guidelines while requiring virtually no human input. Our zero-shot experiments on the clinical named entity recognition benchmarks, 2012 i2b2 EVENT, 2012 i2b2 TIMEX, 2014 i2b2, and 2018 n2c2 showed 25.86%, 4.36%, 0.20%, and 7.75% improvements in strict F1 scores from the no-guideline baseline. The LLM-synthesized guidelines showed equivalent or better performance compared to human-written guidelines by 1.15% to 4.14% in most tasks. In conclusion, this study proposes a novel LLM self-improving method that requires minimal knowledge and human input and is applicable to multiple biomedical domains.

## INTRODUCTION

The use of large language models (LLMs) for clinical information extraction has become increasingly popular in recent years [1]. Methodologies such as few-shot learning, prompt engineering, external knowledge, knowledge distillation, fine-tuning, and Chain-of-Thoughts (CoT) [2–9] have been developed to improve performance, efficiency, robustness, and generalizability. As a subgroup of the prompt engineering methods, the use of annotation guidelines ("annotation guideline-based prompting") is an under-studied field. Annotation guidelines are concisely written documents that provide background knowledge, definitions, and instructions for human annotators to contribute to natural language processing (NLP) projects [10–12]. An important purpose of the guidelines is to define information of interest (e.g., entity and relation types). The definitions in these guidelines are often designed for certain needs and thus unique to specific tasks.

In the general and biomedical NLP, a few previous studies proposed prompting with annotation guidelines written by domain experts as part of the prompt [4,13]. Annotation guideline-based prompting is straightforward, efficient, and consistently delivers high performance. Moreover, as an independent module in prompt engineering, it can be used alongside other methods (e.g., few-shot learning, external knowledge, fine-tuning, CoT). Despite all the benefits, the construction of annotation guidelines is labor-intensive and involves both medical domain knowledge [4] and NLP knowledge [10–12]. Furthermore, its task-specific nature poses challenges to the reusing of guidelines, thereby increasing the effort required for initiating new projects. These are significant obstacles to practical applications.

Fortunately, recent studies have shown the capacity of LLMs for data and knowledge synthesis in which LLMs process and generate data for downstream model training [14–17]. In this study, we propose a novel method that leverages LLMs to synthesize annotation guidelines, which are subsequently used to enhance zero-shot and few-shot performance in information extraction tasks. To our knowledge, this work is the first adoption of the self-improvement framework [18] in the form of synthesized annotation guidelines. Our work has the following contributions: 1) We propose a knowledge-lite method that harvests the knowledge-summarization and text-generation capacity of LLMs while requiring virtually no human input. 2) We generalize the results of previous studies on annotation guideline-based prompting for named entity recognition (NER) [4] to different NLP tasks and biomedical domains. 3) Our ablation study shows both the examples and the narratives in annotation guidelines contribute to information extraction performance.

## RELATED WORK

### Annotation Guideline-based Prompting

In the general NLP domain, a recent study utilized annotation guidelines to enhance zero-shot information extraction [13]. The authors fine-tuned Code-LLaMA [19] to follow guidelines written in Python syntax. In domains such as politics, literature, music, AI, and science, the proposed method showed consistent improvements for zero-shot NER.

In the biomedical domain, a recent study evaluated different prompt engineering methods including annotation guidelines, error analysis, and few-shot learning [4]. The authors inserted a human-written guideline with entity definitions and linguistic rules into the prompt template. On two clinical NER benchmarks (MTSamples and VAERS), the annotation guideline-based method achieved a 3.5% and 7% increase in strict F1 score from baseline.

The annotation guidelines in both studies were written by domain experts for specific tasks. To our knowledge, there is no study utilizing LLMs to synthesize annotation guidelines for information extraction.

**Generative Information Extraction with Large Language Models**

Large language models have been adopted for information extraction in numerous studies [1]. In the biomedical field, popular information extraction tasks include 1) **named entity recognition (NER)**, which aims to extract entities (e.g., medication, treatment, diagnosis) with character spans; 2) **Relation extraction (RE)**, which categorizes relations between named entities (e.g., drug-ADE relation); and 3) **Event extraction (EE)**, which detects medically relevant events (e.g., treatment event, adverse event) and assigns attributes (e.g., timestamp, event type). For NER tasks, several methods including prompt engineering [4], external knowledge [5], knowledge distillation [6,7], fine-tuning [8], and Chain-of-Thoughts (CoT) [9] have been proposed for domains such as medication and adverse drug event (ADE) extraction [8], multilingual clinical concept extraction [6,7], and biomedical entity extraction [4,5]. For RE tasks, few-shot learning [2,3], fine-tuning [20], and data augmentation [21] were evaluated for medication attributes [2], microbiome-disease [20], gene-disease [3,21], and drug-disorder-target relation extractions [21]. For EE tasks, Andrew et al. evaluated different prompt designs for pediatric temporal event extraction using LLM [22]. Ma et al. proposed a framework, DICE, for medical events extraction from public clinical case reports [23].

**Leveraging Large Language Models for Data and Knowledge Synthesis**

Using LLMs to generate data and knowledge for machine learning model development has become an increasingly popular approach. The pre-training of Llama 3 models utilized Llama 2 [24] for training data preparation [14]. Llama 2 was prompted to label the data quality of documents and to annotate code and math text from web data. The labels were used to fine-tune lightweight DistilRoberta classifiers for large-scale training data labeling. A recent study utilized LLMs to synthesize scientific knowledge and perform data inferencing for training downstream machine learning models on 58 tasks across 4 scientific fields [15]. LLMs were prompted to generate domain-specific prediction rules from literature, followed by inferencing datasets to generate features. The features were then used for training interpretable machine learning models (e.g., random forest).

In the biomedical domain, Xu et al. proposed a knowledge-infused prompting method for data augmentation [16]. Entities and relations were extracted from knowledge graphs and generated by LLMs to construct a data-generation prompt. The authors prompted LLMs with it to synthesize training data (i.e., sentences and labels). The training data was then used to fine-tune a downstream PubMedBERT [25] model for 18 biomedical NLP tasks. Wadhwa et al. proposed a Chain-of-thoughts (CoT) approach that prompts GPT-3 to review medical case reports and drug-ADE relation tuples and generate explanations [17]. The gold standard and corresponding explanations were then used to fine-tune a downstream Flan-T5 model for relation extraction tasks.

## METHODS

**Methodology Overview**

We prompt the state-of-the-art open-source LLM, Llama 3.1 405B, to synthesize a task-specific and human-friendly annotation guideline that includes helpful information (e.g., instructions, entity definitions, examples) for performing an NLP annotation task. We then embed the synthesized guideline in a prompt for downstream LLMs to conduct information extraction tasks (Figure 1). Details are described in this section and in the online supplement (Implementation Details section).

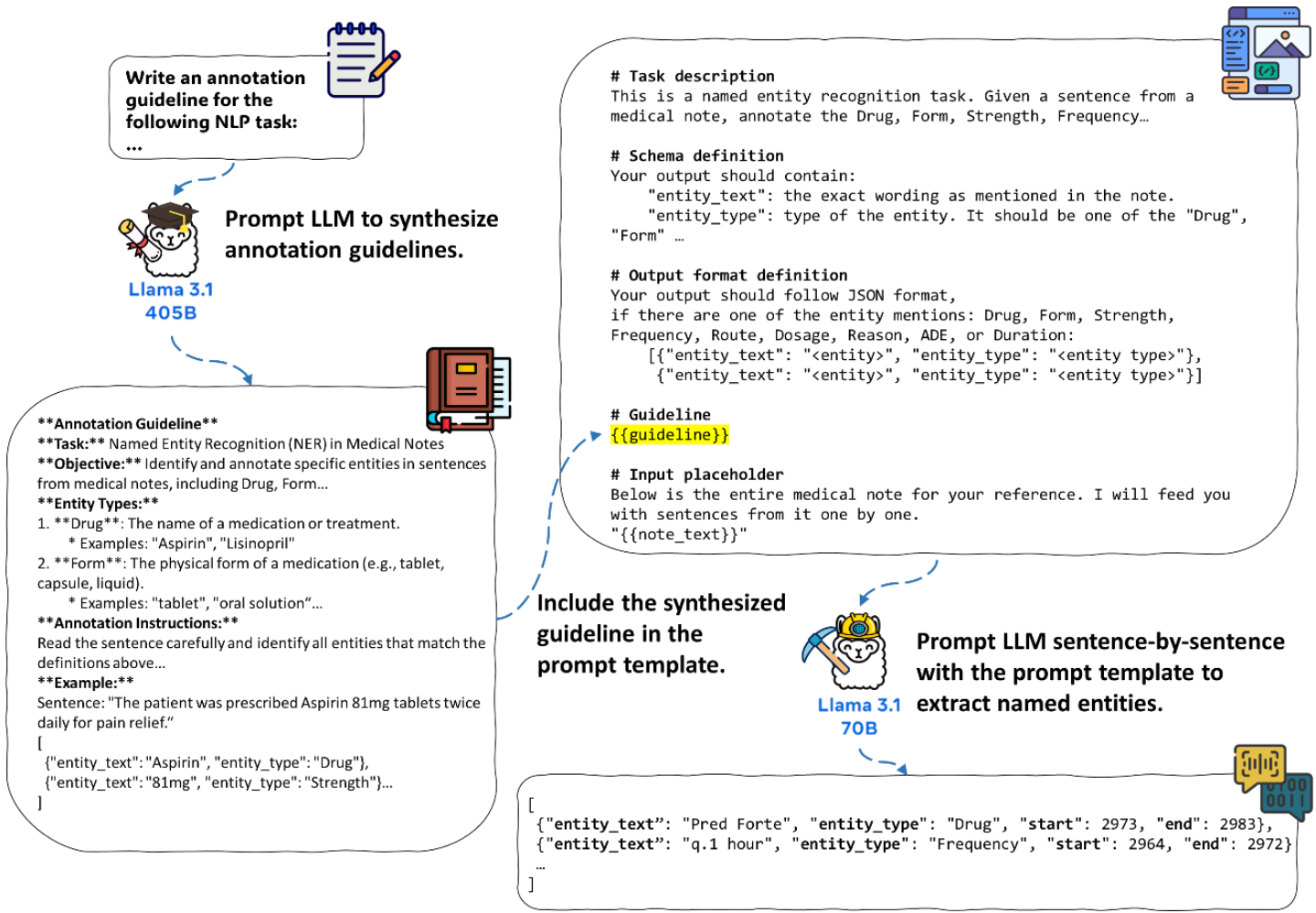

Figure 1: Methodology flowchart for the NER benchmarks. For each benchmark, we prompted Llama 3.1 405B to synthesize an annotation guideline. We proofread and amended it before including it in the prompt template. We then prompt Llama-3.1-70B-Instruct with sentences from clinical notes for named entity extraction. The outputs are JSON lists with entity texts, entity types, and entity spans.

**LLM Synthesis of Annotation Guidelines with Minimal Human Guidance**

We prepare a short instruction ("starter prompt") that includes limited context and virtually no knowledge to prompt the Llama 3.1 405B [14]. For example, the starter prompt for the 2018 n2c2 task is:

*Write an annotation guideline for the following NLP task:*

*This is a named entity recognition task. Given a sentence from a medical note, annotate the Drug, Form, Strength, Frequency, Route, Dosage, Reason, ADE, and Duration.*

*The output should follow JSON format:*

*[{"entity_text": "<entity text>", "entity_type": "<entity type>"}...]*

The LLM synthesizes a comprehensive, human-readable, and domain-specific annotation guideline (denoted as "LLM-synthesized guidelines"). We proofread the LLM's output and apply minor amendments. Table 2 shows the synthesized guideline with amendments. The full text of the starter prompt, LLM-synthesized annotation guidelines, and our amendments are available in supplementary Table S3-S5.

**Datasets and Human-written Annotation Guidelines**

We evaluated our method on three public NLP benchmarks for clinical named entity recognition (NER) and one private dataset for dental adverse event (AE) detection. All datasets have an annotation guideline written by domain experts (denoted as "human-written guidelines"). As a comparison to our proposed method, we used them to prompt LLMs for downstream NLP tasks.

**NER benchmarks**. We adopted the datasets and annotation guidelines from the 2012 [10] and 2014 [11] Integrating Biology and the Bedside (i2b2), and 2018 National NLP Clinical Challenges (n2c2) [12] Natural Language Processing Challenge. For each benchmark, we randomly sampled eight sentences (instead of notes) that included some entities from the training set as the few-shot examples. The rest of the training sets were not used. Summary statistics are available in Table 1. The i2b2/n2c2 annotation guidelines were downloaded from the DBMI portal as PDFs. Despite variation in structure, in general, they include the following sections: 1) an introduction that explains the motivation of the work and provides a context for the annotation task; 2) an itemized definition of each entity type, often with examples for clarification; 3) Additional desiderata provided in the form of notes or FAQs. We extracted the relevant sections into plain text format (Table S2).

**Adverse event dataset**. An electronic health records (EHRs) from a large academic dental school which is part of the BigMouth [26] network includes 1,530,876 dental notes of 134,889 patients from 2014 to 2024. We randomly sampled 500 notes for development and evaluation. Our dentist co-author, KZ, manually labeled dental adverse events [27,28] as a binary flag at the note level. 42 notes were excluded due to insufficient information for determining adverse events. Subsequently, 8 notes (4 notes with AEs and 4 notes without AEs) were sampled as few-shot examples, while the remaining 450 notes were used as the test set. Among the test set, 37 (8.2%) notes included some AEs (Table 1). KZ prepared a two-paged annotation guideline with general comments, a codebook, and references. Adverse events were defined by six categories (i.e., hard tissue damage, infection, soft tissue injury, pain, nerve damage, and others) with subcategories and descriptions (Table S2)).

**Prompt Template**

For the downstream LLMs to intake the annotation guidelines, we designed task-specific prompt templates to embed them. Adopted from previous studies [2,4,29], our prompt templates are comprised of five sections: 1) **task description,** which provides the context and lists the entity types; 2) **schema definition,** which explains the output tags: "entity_text" and "entity_type;"; 3) **output format definition,** which regulates the LLM outputs as JSON format; 4) **guideline**, where we place the human-written guidelines or LLM-synthesized guidelines; 5) **input placeholder,** where we place the entire note text. For experiments involving few-shot examples, there is an additional **Examples** section where few-shot examples are provided. The full text of prompt templates is available in Table S1.

*Table 1: Summary of datasets and annotation guidelines*

|  | 2012 i2b2 | 2014 i2b2 | 2018 n2c2 |
|---|---|---|---|
| **Clinical notes** | | | |
| Number of notes in the test set | 120 | 514 | 202 |
| Total number of sentences | 4,782 | 23,563 | 28,718 |
| Total number of entities | 15,376 | 11,462 | 32,822 |
| Entities per sentence, median [Q1, Q3] | 3.0 [2.0,4.0] | 0.0 [0.0,0.0] | 0.0 [0.0,1.0] |
| Entities per sentence, mean (Std Dev) | 3.22 (2.58) | 0.49 (2.01) | 1.14 (3.71) |
| Entities per note, median [Q1, Q3] | 115.0 [73.75,162.25] | 18.5 [13,28] | 143.5 [98,212] |
| **Annotation guideline** | | | |
| Pages | 40 | 2 | 13 |
| Total word count | 13,414 | 482 | 3,213 |
| Words used in the prompt | 3,754 | 472 | 1,724 |
| | **Dental Adverse Event** | | |
| **Clinical notes** | | | |
| Number of notes | 450 | | |
| Number of notes with some AEs | 37 | | |
| **Annotation guideline** | | | |
| Pages | 2 | | |

| Total word count | 832 |
|---|---|
| Words used in the prompt | 744 |
| Summary statistics only represent the test set of each dataset. ||

Table 2: LLM-synthesized annotation guideline for the 2018 n2c2

**2018 (Track 2) ADE and Medication Extraction Challenge**

**Annotation Guideline**

**Task:** Named Entity Recognition (NER) in Medical Notes

**Objective:** Identify and annotate specific entities in sentences from medical notes, including Drug, Form, Strength, Frequency, Route, Dosage, Reason, ADE (Adverse Drug Event),
and Duration.

**Entity Types:**

1. **Drug**: The name of a medication or treatment.
    * Examples: "Aspirin", "Lisinopril"
2. **Form**: The physical form of a medication (e.g., tablet, capsule, liquid).
    * Examples: "tablet", "oral solution"
3. **Strength**: The strength of a medication.
    * Examples: the "20mg" in the sentence, "Patient was given 1 Prednisone 20mg tablet". Note that the "1" in the sentence should not be annotated.
4. **Frequency**: How often a medication is taken.
    * Examples: "twice daily", "every 4 hours"
5. **Route**: The method by which a medication is administered (e.g., oral, intravenous).
    * Examples: "PO" (oral), "IV" (intravenous)
6. **Dosage**: The dosage of medication.
    * Examples: The "1" in the sentence, "Patient was given 1 Prednisone 20mg tablet". Note that the "20mg" should not be annotated.
7. **Reason**: The medical condition or symptom being treated with a medication.
    * Examples: "hypertension", "pain relief"
8. **ADE** (Adverse Drug Event): An unwanted effect caused by a medication.
    * Examples: "rash", "nausea"
9. **Duration**: The length of time a medication is taken.
    * Examples: "2 weeks"

**Annotation Instructions:**

1. Read the sentence carefully and identify all entities that match the definitions above.
2. For each entity, create a separate annotation in JSON format:
```json
[{"entity_text": "<entity text>", "entity_type": "<entity type>"}...]
```
3. Replace `<entity text>` with the exact text of the entity as it appears in the sentence.
4. Replace `<entity type>` with one of the nine entity types listed above (e.g., "Drug", "Form", etc.).
5. If an entity has multiple words, annotate the entire phrase (e.g., "blood pressure medication" would be annotated as a single entity).
6. If an entity is implied but not explicitly stated, do not annotate it.

**Example:**

Sentence: "The patient was prescribed 1 x Aspirin 81mg tablets twice daily for pain relief."

Annotation:
```json
[
 {"entity_text": "Aspirin", "entity_type": "Drug"},
 {"entity_text": "1", "entity_type": "Dosage"},
```

```
  {"entity_text": "81mg", "entity_type": "Strength"},
  {"entity_text": "tablets", "entity_type": "Form"},
  {"entity_text": "twice daily", "entity_type": "Frequency"},
  {"entity_text": "pain relief", "entity_type": "Reason"}
]
```
By following these guidelines, you will help create a high-quality dataset for training and evaluating NLP models in the medical domain.

**Generative Information Extraction with Large Language Models**

We chose the relatively lightweight 70-billion-parameter Llama 3.1 ("Meta-Llama-3.1-70B-Instruct") for a balance between performance and computational cost. It saves 82.7% of GPU memory and float point operations while achieving 90% performance on general knowledge (MMLU-Pro benchmark) [30] compared to the 405-billion version [14].

For all information extraction tasks, we used a system prompt:

> "You are a highly skilled clinical AI assistant, proficient in reviewing clinical notes and performing accurate information extraction".

For NER tasks, we split the notes into sentences and prompted the LLM to extract named entities by generating them in JSON format. We then used regular expressions to identify the entity spans. LLM inference was performed with the vLLM inference engine [31]. The computation was performed on a server with eight NVIDIA A100 GPUs.

For the Dental adverse event detection, we prompted the LLM with the entire note text to generate adverse events and categories as a JSON list. The LLM inference was performed with the Ollama inference engine [32]. The computation was performed on a server with four NVIDIA V100 GPUs.

**Evaluation**

We adopted the official i2b2/nn2c2 Python grading scripts and reported the primary metrics. We evaluated the dental adverse event detection task as document-level classification (i.e., include/ not include the adverse event(s)) and reported the precision, recall, F1 score, F2 score, and accuracy. Since adverse events are rare yet have a significant impact on patient safety and quality of care, this NLP task emphasizes recall over precision.

**Ablation Study**

NLP guidelines generally contain both declarative instructions (defining the task in a narrative manner) and exemplative instructions (examples with correct or incorrect annotations) It is important to understand which of these approaches of a guideline the LLM relies upon in order to focus future efforts on the use of guideline in LLM-based information extraction. To this end, we manually excluded examples from the human-written guidelines from 2012 i2b2 and 2018 n2c2 before prompting LLMs. We did not include the 2014 i2b2 and dental adverse event task in this ablation study since there are not many examples provided in those guidelines.

## RESULTS

**Qualitative Analysis of the LLM-synthesized Annotation Guidelines**

Llama 3.1 405B automatically synthesized the annotation guidelines in Markdown language including header tags, bold fonts, item lists, and code blocks (e.g., "```json```"). The guideline had a total of 440, 512, 350, and 339 words for the 2012 i2b2, 2014 i2b2, 2018 n2c2, and dental adverse event detection tasks, respectively (Table S4). Without specific instructions in the starter prompts, the synthesized annotation guidelines follow a consistent pattern with six sections: 1) **Title**, (e.g., "Annotation Guideline for Adverse Events in Dental Notes", "Annotation Guidelines for Protected Health Information (PHI) Named Entity Recognition"); 2) **Introduction**, which provides a context to the annotation task (e.g., "Task Overview", "Introduction", "Task)"); 3) **Annotation types**, which defines the annotation types and provides short examples. This section is titled "Entity types" for all the three NER guidelines and "Adverse Event Categories" for the dental adverse event detection guideline; 4) **Instructions**, which defines the output format and provides additional guidance such as "If an entity is mentioned multiple times in the same context, annotate each occurrence separately"(e.g., "JSON

Output Format", "Annotation Instructions"); 5) **Example**, where synthesized notes or sentences are provided with the expected outputs following the required JSON format; and 6) **Quality Control**, where hints and reminders are listed (e.g., "Review your annotations carefully to ensure accuracy and consistency.", "Choose the most specific category label that applies to the adverse event."). This section is titled "Quality Control" in the adverse event detection and 2014 i2b2 guidelines and titled "Note" in the 2012 i2b2 guideline, while omitted in the 2018 n2c2.

**Proofreading and Amendments**

We manually proofread the LLM-synthesized annotation guidelines. We found the Markdown syntax was completely correct. We did not notice any grammar or spelling errors. However, looking into the contents, especially the entity types section, we found some inconsistencies. In the 2012 i2b2, the LLM created unintended subtypes: test, problems, and treatments for the EVENT, and subtypes: date, time, duration, and frequency for the TIMEX. We removed those from the synthesized guideline. In the 2014 i2b2, we provided additional hints that prefixes (e.g., Mr. Ms.) and titles (e.g., MD.) should not be annotated. We also expanded the definition of age to include both patients and families. We found that LLM restricted the "Date" entity to patients' medical records or treatment dates. We expanded it to any date mentioned. In the 2018 n2c2, we found that LLM confused the "Strength" and the "Dosage" entity types. We corrected it and provided an example to highlight the differences. In the dental adverse event task, we emphasized that AEs are "physical harm associated with dental treatment". We also specified the distinguishment between adverse events, errors, and deviations from standards of care (Table S5).

The overlap coefficient (see formula in supplementary) between the raw syntheses and the amended versions are 95.76%, 99.76%, 99.53%, and 99.96% for the 2012 i2b2, 2014 i2b2, 2018 n2c2, and dental adverse event, respectively.

**Information Extraction Performance**

**Named entity recognition (NER) benchmarks**. In zero-shot prompting, when the amended LLM-synthesized guidelines were used, there was a consistent improvement from the baseline (no guideline) by 25.86%, 4.36%, 0.2%, and 7.75% for the 2012 i2b2 EVEN, 2012 i2b2 TIMEX, 2014 i2b2, and 2018 n2c2, respectively. When the raw LLM-synthesized guidelines were used, the 2012 i2b2 EVENT showed an obvious improvement by 18.3%, while the 2012 i2b2 TIMEX and 2014 i2b2 showed slightly decreased F1 scores by 2.48% and 0.48%, respectively. The human-written guidelines resulted in a consistent increase of 2.94% to 21.72% in F1 scores. Interestingly, in the 2012 i2b2 EVENT, 2012 i2b2 TIMEX, and 2018 n2c2, the amended LLM-synthesized guidelines resulted in slightly better performance than the human-written guidelines by 4.14%, 1.42%, and 1.15%, respectively. In few-shot prompting, the amended LLM-synthesized guidelines showed benefits in most benchmarks by 0.49% to 2.94%, except for the 2014 i2b2 which had a 1.1% decrease. The raw LLM-synthesized guidelines showed similar performance as the baseline. While the human-written guidelines resulted in performance gains in most benchmarks except for the 2012 i2b2 EVENT (Figure 2, Table 3).

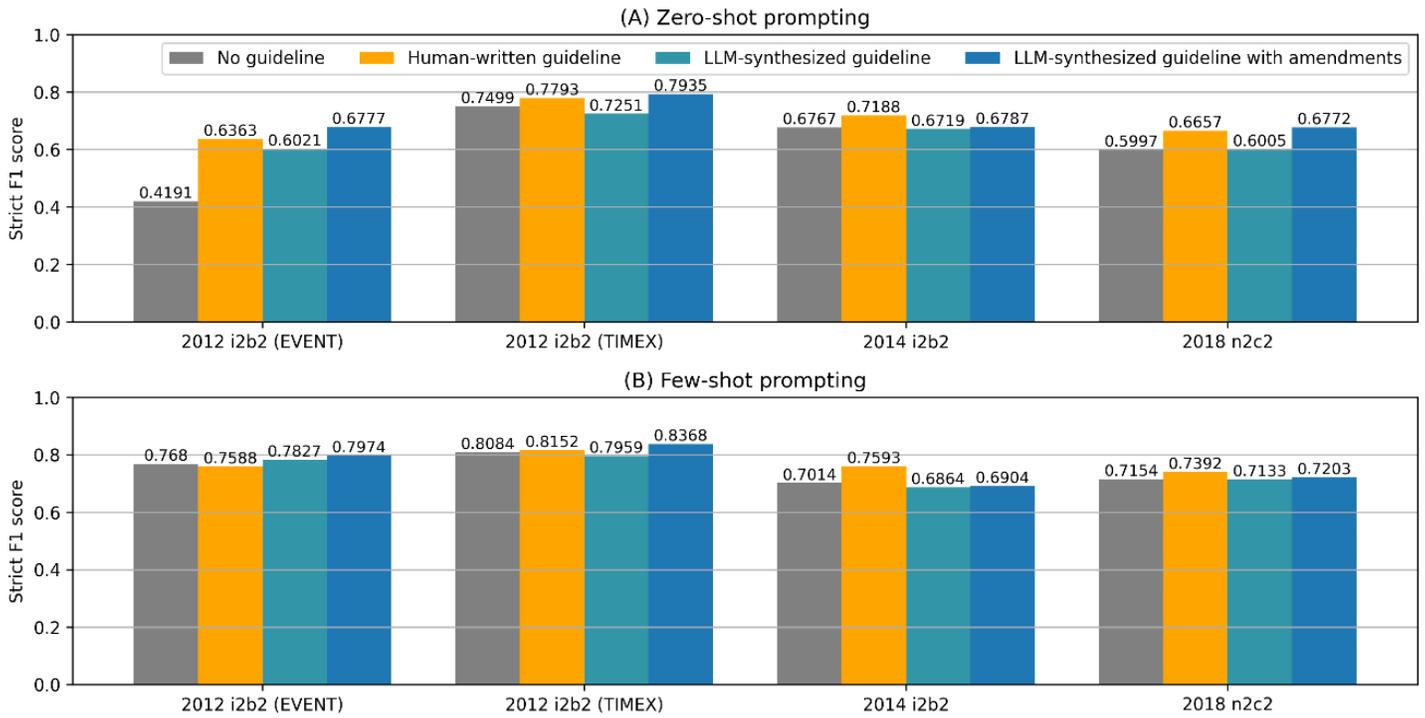

*Figure 2: Named entity recognition performances with human-written guidelines, LLM-synthesized guidelines, and LLM-synthesized guidelines with amendments. (A) The zero-shot prompting and (B) the few-shot prompting.*

**Dental adverse event detection dataset.** In zero-shot prompting, the amended LLM-synthesized guidelines resulted in an improvement in the F1 score by 1.04%. The raw LLM-synthesized guidelines showed a slight increase in F1 score by 0.33% and an increase in the recall by 2.7%. The human-written guidelines had an obvious benefit on the F1 score by 19.26%. In few-shot prompting, both the raw LLM-synthesized guidelines and the amended version resulted in lower F1 scores by 8.11% and 7.67% respectively. However, the recalls increased by 10.81%. The human-written guideline showed a consistent increase in both F1 score (5.7%) and recall (2.7%) (Figure 3, Table 3).

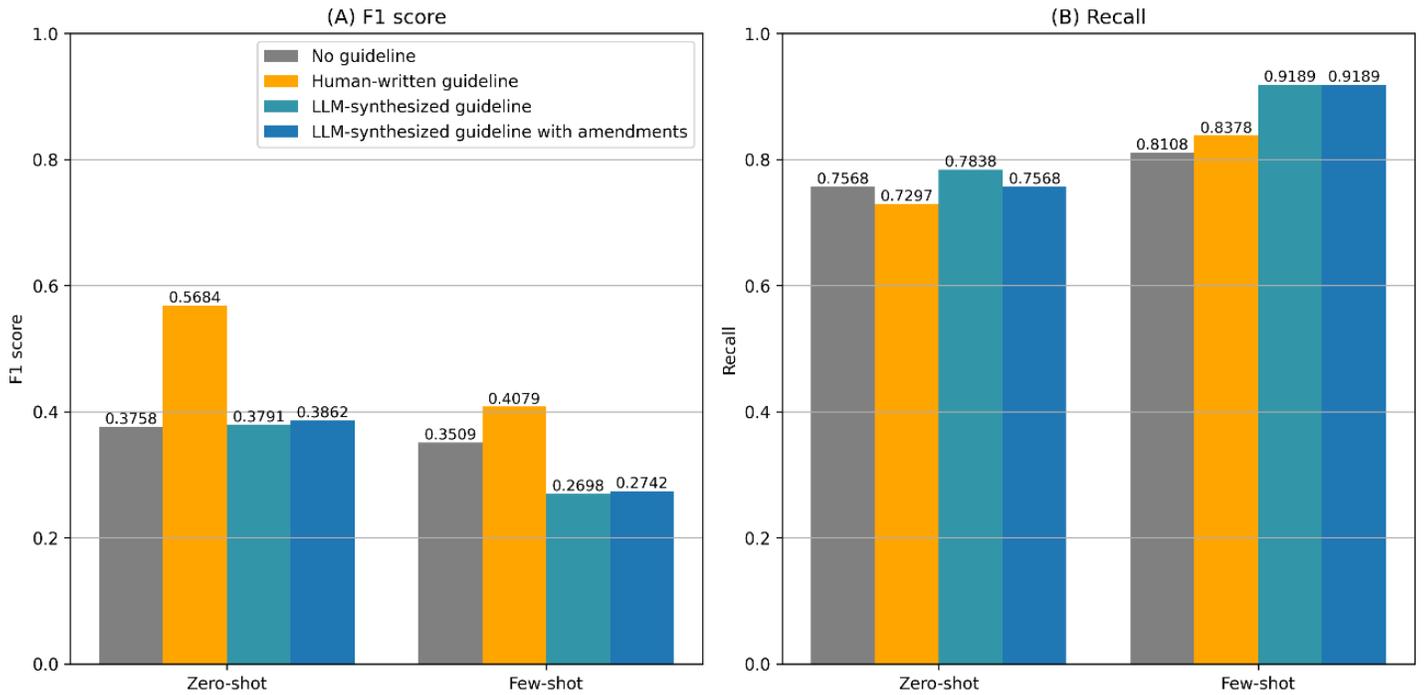

*Figure 3: Dental adverse event detection with human-written guidelines, LLM-synthesized guidelines, and LLM-synthesized guidelines with amendments. (A) the F1 scores and (B) recall.*

*Table 3: Information extraction performance with human-written guidelines, LLM-synthesized guidelines, and LLM-synthesized guidelines with amendments.*

| | **2012 Temporal Relations Challenge** | | | | | | |
|---|---|---|---|---|---|---|---|
| Experiments | | EVENT | | | TIMEX | | |
| | | Precision | Recall | F1 | Precision | Recall | F1 |
| Zero-shot | No guideline | 0.9469 | 0.2691 | 0.4191 | 0.7438 | 0.756 | 0.7499 |
| | Human-written GL | 0.8679 | 0.5023 | 0.6363 | 0.7121 | 0.8604 | 0.7793 |
| | LLM-synthesized GL | 0.9489 | 0.441 | 0.6021 | 0.839 | 0.6385 | 0.7251 |
| | LLM-synthesized GL with amendment | 0.9656 | 0.5221 | 0.6777 | 0.8423 | 0.75 | 0.7935 |
| Few-shot | No guideline | 0.8441 | 0.7044 | 0.768 | 0.8752 | 0.7511 | 0.8084 |
| | Human-written GL | 0.8049 | 0.7177 | 0.7588 | 0.865 | 0.7709 | 0.8152 |
| | LLM-synthesized GL | 0.8938 | 0.6962 | 0.7827 | 0.852 | 0.7467 | 0.7959 |
| | LLM-synthesized GL with amendment | 0.881 | 0.7283 | 0.7974 | 0.9001 | 0.7819 | 0.8368 |
| **2014 De-identification Challenge** | | | | | | | |
| Experiments | | Strict | | | Relaxed | | |
| | | Precision | Recall | F1 | Precision | Recall | F1 |
| Zero-shot | No guideline | 0.6925 | 0.6617 | 0.6767 | 0.6948 | 0.6638 | 0.679 |
| | Human-written GL | 0.7649 | 0.678 | 0.7188 | 0.7672 | 0.6801 | 0.721 |
| | LLM-synthesized GL | 0.6738 | 0.67 | 0.6719 | 0.6762 | 0.6724 | 0.6743 |
| | LLM-synthesized GL with amendment | 0.6725 | 0.6851 | 0.6787 | 0.6746 | 0.6873 | 0.6809 |
| Few-shot | No guideline | 0.6683 | 0.7379 | 0.7014 | 0.6703 | 0.7401 | 0.7035 |
| | Human-written GL | 0.7704 | 0.7486 | 0.7593 | 0.7722 | 0.7504 | 0.7611 |
| | LLM-synthesized GL | 0.6442 | 0.7347 | 0.6864 | 0.6461 | 0.7369 | 0.6885 |

|  |  | Precision | Recall | F1 | Precision | Recall | F1 |
|---|---|---|---|---|---|---|---|
|  | LLM-synthesized GL with amendment | 0.6458 | 0.7416 | 0.6904 | 0.6478 | 0.7438 | 0.6925 |

**2018 (Track 2) ADE and Medication Extraction Challenge**

| | Experiments | Strict | | | Lenient | | |
|---|---|---|---|---|---|---|---|
| | | Precision | Recall | F1 | Precision | Recall | F1 |
| Zero-shot | No guideline | 0.7495 | 0.4998 | 0.5997 | 0.9061 | 0.6016 | 0.7231 |
| | Human-written GL | 0.7872 | 0.5767 | 0.6657 | 0.9282 | 0.6774 | 0.7832 |
| | LLM-synthesized GL | 0.7041 | 0.5234 | 0.6005 | 0.8501 | 0.6293 | 0.7232 |
| | LLM-synthesized GL with amendment | 0.783 | 0.5967 | 0.6772 | 0.8825 | 0.6695 | 0.7614 |
| Few-shot | No guideline | 0.852 | 0.6166 | 0.7154 | 0.963 | 0.692 | 0.8053 |
| | Human-written GL | 0.8559 | 0.6506 | 0.7392 | 0.954 | 0.7227 | 0.8224 |
| | LLM-synthesized GL | 0.8055 | 0.64 | 0.7133 | 0.9104 | 0.7191 | 0.8035 |
| | LLM-synthesized GL with amendment | 0.8206 | 0.6418 | 0.7203 | 0.9207 | 0.7166 | 0.8059 |

**Dental Adverse Event Detection**

| | Experiments | Precision | Recall | F1 | F2 | Accuracy |
|---|---|---|---|---|---|---|
| Zero-shot | No guideline | 0.25 | 0.7568 | 0.3758 | 0.5385 | 0.7933 |
| | Human-written GL | 0.4655 | 0.7297 | 0.5684 | 0.6553 | 0.9089 |
| | LLM-synthesized GL | 0.25 | 0.7838 | 0.3791 | 0.5492 | 0.7889 |
| | LLM-synthesized GL with amendment | 0.2593 | 0.7568 | 0.3862 | 0.5469 | 0.8022 |
| Few-shot | No guideline | 0.2239 | 0.8108 | 0.3509 | 0.5319 | 0.7533 |
| | Human-written GL | 0.2696 | 0.8378 | 0.4079 | 0.5894 | 0.8 |
| | LLM-synthesized GL | 0.1581 | 0.9189 | 0.2698 | 0.4683 | 0.5911 |
| | LLM-synthesized GL with amendment | 0.1611 | 0.9189 | 0.2742 | 0.4735 | 0.6 |

**Ablation Study**

The human-written guidelines without examples resulted in improvements of 23.51%, 2.13%, and 3.22% from baseline for the 2012 i2b2 EVENT, 2012 i2b2 TIMEX, and 2018 n2c2. Compared to the version with examples, there was a slight improvement in the 2012 i2b2 (1.79%) and a decrease in 2012 i2b2 TIMEX (0.81%) and 2018 n2c2 (3.38%) (Figure 4).

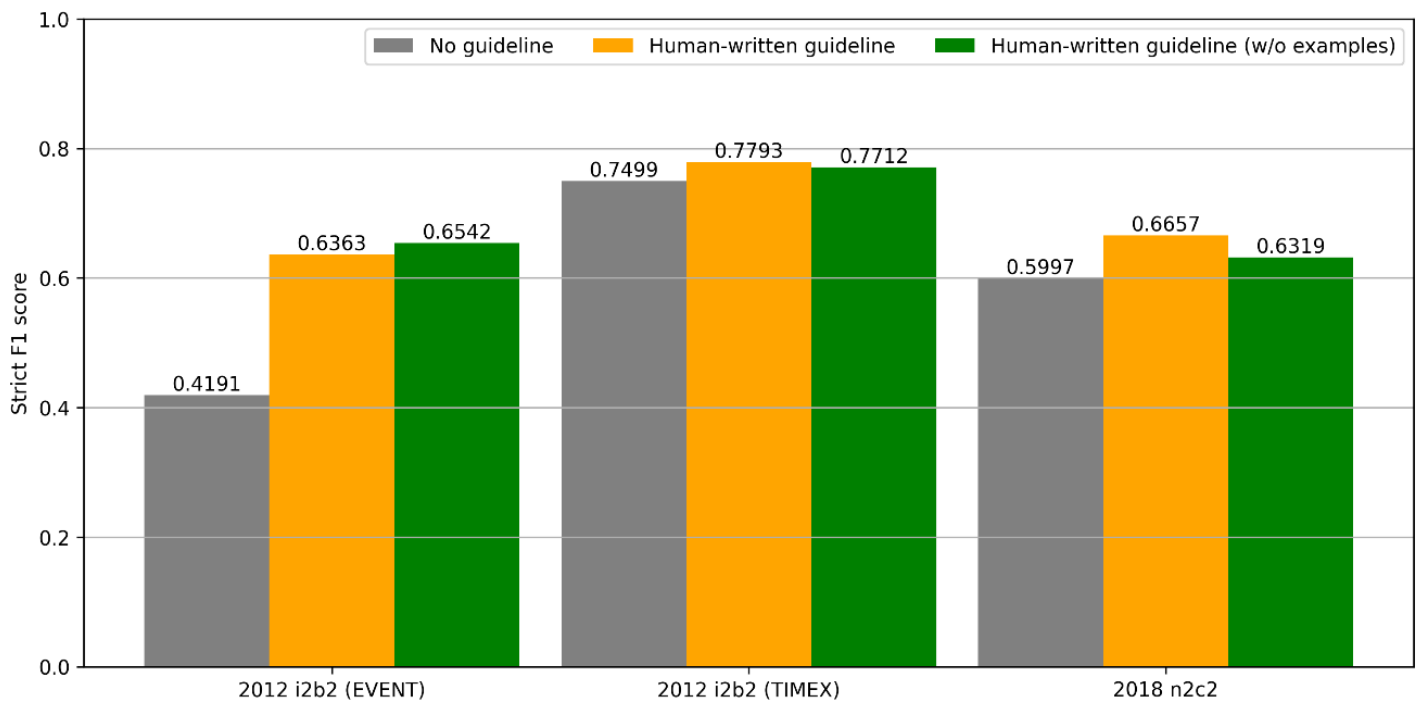

*Figure 4: Ablation study comparing named entity recognition performances of human-written guidelines with and without examples.*

## DISCUSSION

In this study, we proposed a knowledge-lite method for clinical information extraction that requires virtually no human input while resulting in consistent performance gain in zero-shot and few-shot clinical information extraction. With a short starter prompt and a few amendments (less than 5% changed), our proposed method achieved a 0.2% to 25.86% improvement in strict F1 score for zero-shot NER. Our method resulted in a 1.04% improvement in the F1 score for the dental adverse event detection dataset. For few-shot prompting, our method achieved 0.49% to 2.94% improvement in most NER benchmarks. While showing a decreased F1 score on the dental adverse event detection dataset, the recall increased by over 10%, which is preferable given the purpose of adverse event detection. Interestingly, in most NER benchmarks, our method resulted in slightly better performance than the human-written guidelines by 1.15% to 4.14%. This result indicates that using LLM to synthesize guidelines is not only time-saving and knowledge-free, but could also improve accuracy in some cases. Our experiments involved various clinical NLP tasks across distinct domains. We believe our method is generalizable to multiple biomedical domains.

The Llama 3.1 405B-synthesized high-quality annotation guidelines followed a consistent pattern (i.e., title, introduction, entity types, instruction, examples, and quality control) without human guidance. This result suggests that the LLM has established semantic embedding of the concept of "annotation guideline" during the pre-training and can generate informative, human-friendly, and task-focused annotation guidelines.

An annotation guideline often provides context and medical domain knowledge, which contributes to the performance of information extraction. But more importantly, it provides definitions that are specific to the design of the study. For example, the "TIMEX" entity in the 2012 i2b2 is defined as temporal expressions including date, time, duration, and frequency. This definition serves the purpose of capturing the temporal information of clinical events [10]. A conceptually similar entity, "Date", in the 2014 i2b2 however, is defined as "Any calendar date, including years, seasons, months, and holidays, while not including the time of day" to serve the purpose of PHI de-identification [11]. Such definitions are project and purpose dependent which can only be completed by the researchers. Therefore, it is necessary for human researchers to proofread and amend the LLM-synthesized guidelines. As a future research direction, we will explore an iterative guideline development method in which we 1) prompt LLM to synthesize guidelines, 2) human proofread and amend, 3) LLM revise guidelines, and 4) repeat steps 2 and 3 until the performance on the development set stops improving.

Previous studies prompted LLMs with annotation guidelines that include both narrative and embedded examples [4]. Such study design makes it difficult to distinguish the contributions from the two components. An extreme hypothesis is that LLMs learn solely from the examples and not the narrative of the guidelines. The guideline-based prompting method would then be equivalent to few-shot prompting. To rule out such a hypothesis and to quantify the contribution of the narrative and example components, our ablation study excluded the embedded examples. The result shows an overall improvement from the baseline (no guideline) by 2.13% to 23.51%, indicating that the narrative contributes to the performance. Compared to the full guidelines, the ablation resulted in minor decreases in most benchmarks. This indicates that both the narrative task descriptions and embedded examples contribute to the performance.

Despite the promising results, this study has a few limitations. First, we only evaluated Llama-3.1, while other highly-rated LLMs, including GPT-4 and Claude 3, were not evaluated. Due to the data user agreement (DUA) of the i2b2/ n2c2 datasets, uploading the clinical notes to external API servers is prohibited. Our IRB for the dental adverse event dataset requires the dental notes to stay within the organization. Therefore, we only consider open-source LLMs that can be deployed on-premise.Llama-3.1 has high reported performance on multiple NLP benchmarks [14]. We use it to represent the state-of-the-art open-source LLMs at this moment. Secondly, in the ablation study, ideally, there should be a comparison of three settings: 1) full guideline, 2) narrative only, 3) and example only. We did not evaluate the example-only setting. This is due to the writing styles in the existing guidelines. Most examples are written in the context of the narrative. Isolating the examples would make them meaningless. To perform such an in-depth analysis, the annotation guidelines need to be re-designed.

## CONCLUSIONS

We proposed a novel method that prompts LLMs to generate annotation guidelines for information extraction and achieved consistent improvements in zero-shot and few-shot prompting. Our method requires minimal knowledge and human input and is applicable to multiple biomedical domains.